\documentclass[twocolumn]{fairmeta}

\usepackage{amsmath,amsfonts,bm}

\def\eqref#1{equation~\ref{#1}}

\def\1{\bm{1}}

\DeclareMathAlphabet{\mathsfit}{\encodingdefault}{\sfdefault}{m}{sl}
\SetMathAlphabet{\mathsfit}{bold}{\encodingdefault}{\sfdefault}{bx}{n}

\usepackage{float}
\usepackage{graphicx}
\graphicspath{{figs/}}
\usepackage{multirow}
\usepackage[utf8]{inputenc} 
\usepackage[T1]{fontenc}    
\usepackage{hyperref}       
\usepackage{url}            
\usepackage{booktabs}       
\usepackage{amsfonts}       
\usepackage{nicefrac}       
\usepackage{microtype}      
\usepackage{xcolor}         
\usepackage{tablefootnote} 
\usepackage{acronym}
\usepackage{soul}
\definecolor{eeg}{RGB}{51,111,162}  
\definecolor{meg}{RGB}{58,146,58}  
\definecolor{fmri3t}{RGB}{180,46,50}  
\definecolor{fmri7t}{RGB}{223,193,32}  

\title{Disentangling the Factors of Convergence between Brains and Computer Vision Models}

\author[1,2]{Joséphine Raugel}
\author[1]{Marc Szafraniec}
\author[1]{Huy V. Vo}
\author[1]{Camille Couprie}
\author[1]{Patrick Labatut}
\author[1]{Piotr Bojanowski}
\author[2]{Valentin Wyart}
\author[1]{Jean-Rémi King}

\affiliation[1]{Meta AI}
\affiliation[2]{Ecole Normale Supérieure - PSL Université}

\correspondence{\email{\{josephiner,jeanremi\}@meta.com}}

\abstract{
Many AI models trained on natural images develop representations that resemble those of the human brain. However, the exact factors that drive this brain-model similarity remain poorly understood. In order to disentangle how the model architecture, training recipe and data type independently lead a neural network to develop brain-like representations, we trained a family of self-supervised vision transformers (DINOv3) that systematically varied these different factors. We compare their representations of natural images to those of the human brain recorded with both ultra-high field functional magnetic resonance imaging (fMRI) and magneto-encephalography (MEG), providing high resolution in spatial and temporal analyses. We assess the brain-model similarity with three complementary metrics focusing on overall representational similarity, topographical organization, and temporal dynamics. We show that all three factors -- model size, training amount, and image type -- independently and interactively impact each of these brain similarity metrics. In particular, the largest DINOv3 models trained with the largest amount of human-centric images reach the highest brain-similarity scores. Importantly, this emergence of brain-like representations in AI models follows a specific chronology during training: models first align with the early representations of the sensory cortices, and only align with the late and prefrontal representations of the brain with considerably more training data. Finally, this developmental trajectory is indexed by both structural and functional properties of the human cortex: the representations that are acquired last by the models specifically align with the cortical areas with the largest developmental expansion, the largest thickness, the least myelination, and the slowest timescales. Overall, these findings disentangle the interplay between architecture and experience in shaping how artificial neural networks come to see the world as humans do, thus offering a promising framework to understand how the human brain comes to represent its visual world.
}

\begin{document}

\twocolumn[%
  \mymaketitle
    \begin{center}
    \includegraphics[width=1\linewidth]{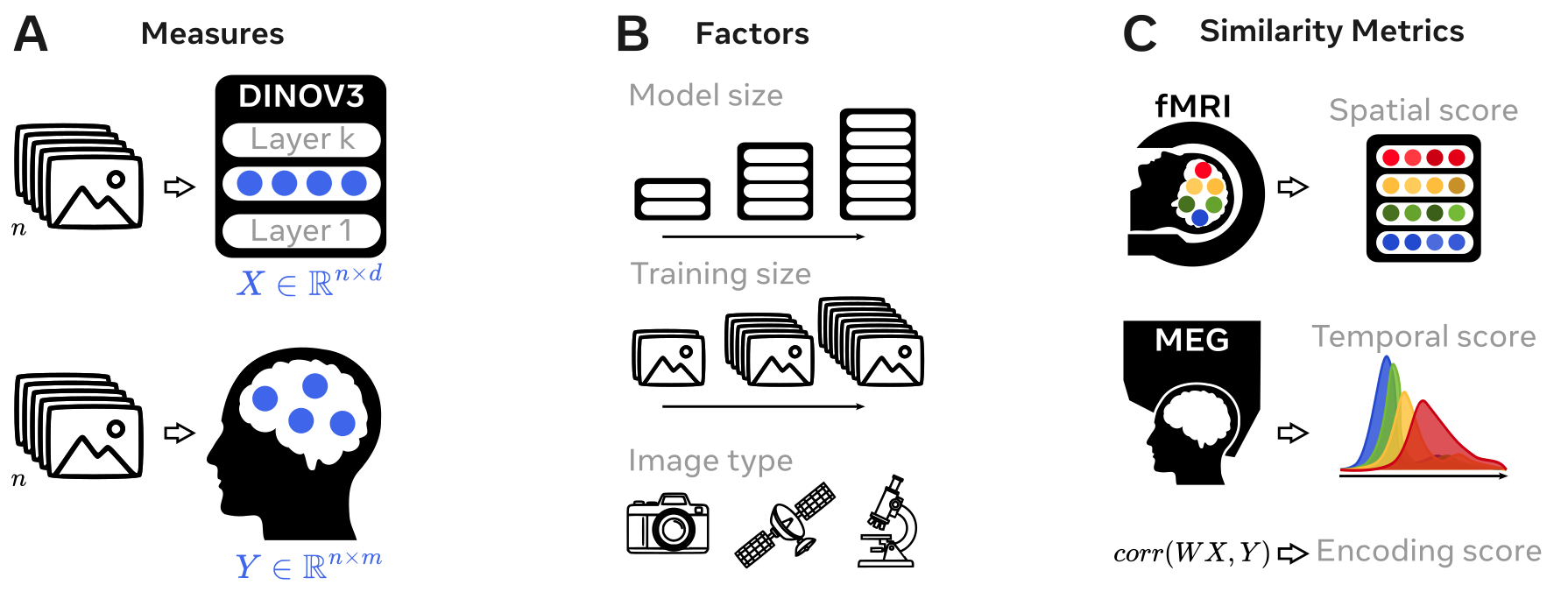}
        \captionsetup{labelformat=empty}
    \begin{minipage}{\textwidth}
    \captionof{figure}{\label{fig:1} \small
    \textbf{A.} We compare the activation of DINOv3, a state-of-the-art self-supervised computer vision model trained on natural images, to the activations of the human brain in response to the same images. 
    \textbf{B.} To understand the factors that make DINOv3 more-or-less similar to the brain, we train \textit{from scratch} a variety of models on different image domains (pictures from human-centric cameras, satellite images or biological data), and with a varying amount of data.
    \textbf{C.} We compare each model to both functional Magnetic Resonance Imaging (fMRI, with high spatial resolution) and Magneto-Encephalography (MEG, with high temporal resolution) by computing the overall linear similarity of their representations (encoding score) and the similarity of their hierarchical organization (spatial and temporal scores).}
\end{minipage}
    \label{fig:methods}

    \end{center}

]

\section{Introduction}
\paragraph{Brain-AI similarity.} Deep learning has transformed computer vision over the past decade. 
State-of-the-art deepnets now achieve human-level or superior performance across a variety of tasks including classification \citep{dinov3,tschannen2025siglip2multilingualvisionlanguage}, object detection \citep{redmon2016lookonceunifiedrealtime}, semantic segmentation \citep{cheng2021mask2former}, and medical image analysis \citep{esteva2017dermatologist,cellular_dino}. 

Surprisingly, the internal representations of these deep learning models appear to be related to those of the human brain: multiple electrophysiology \citep{Yamins2014, Yamins2016, schrimpf2018, Zhuang2021}, functional Magnetic Resonance Imaging \citep{eickenberg2017seeing, millet2023realisticmodelspeechprocessing, Doerig2025, tang2023brainencodingmodelsbased, nikolaus2024modalityagnosticfmridecodingvision}, magneto-encephalography studies \citep{Cichy2016, Seeliger2018, caucheteux2022brains, banville2025scaling} have now consistently shown that the activation patterns of these models linearly map onto those of the cortex in response to the same images.

\paragraph{Theoretical importance.} Understanding the principles at the origin of this representational similarity between AI models and the human brain is of primary importance, to understand the laws of information processing that may be universally shared across neural networks. Indeed, several lines of research \citep{Hasson2020, platonic_hypothesis, vanrossem2024representationsalignuniversalityrepresentation, Cagnetta2024, Mehrer2020, Mahner2025, simkova2025representationsvisionlanguageconverge} suggest that there exists universal principles that constrain the structure and emergence of representations in neural networks.

\paragraph{Challenge: Unclear causes.} However, the precise factors responsible for the representational similarity between computer vision models and the human remain currently unclear. This gap of knowledge is partly due to the fact that previous studies primarily focused on pretrained networks, that \emph{simultaneously} vary in training objectives, architectures and data regime \citep{conwell2021what, rajesh2024brainlikeemergentpropertiesdeep}. How each of these factors independently and interactively lead a model to converge to brain-like representations thus remains unclear. 

To address this issue, we systematically train a variety of DINOv3 models \citep{dinov3}, while independently varying their size, data type and training quantity. DINOv3 has the advantage of being self supervised, and can thus be trained on different types of naturalistic but non-human centric and non-labelled data such as satellite images \citep{dinov3} and biological images \citep{cellular_dino}.

Here, we compare a variety of DINOv3 models to the brain responses to images, as recorded with ultra high field (7T) functional MRI and magneto-encephalography (MEG) to get a high spatial and temporal resolution of the cortical representations, respectively. For this, we implement three similarity metrics. First, we use a standard linear mapping metric, often referred to as \emph{encoding score} \citep{naselaris2011encoding}, which evaluates the linear correspondence between the representations of two systems. Second, we evaluate, with fMRI, whether this linear mapping follows a similar \emph{spatial} organization, whereby the first and last layers of the model would best match the sensory visual and prefrontal cortices, respectively. Finally we evaluate, with MEG, whether this mapping follows a similar \emph{temporal} organization, whereby the first and last layers of the model best match the early and late MEG responses, respectively.

\section{Method}
\subsection{Approach}
We aim to identify the factors that make modern computer vision models process and represent natural images similarly to the human brain.
Following previous work~\citep{kriegeskorte,dicarlo,king_dehaene}, we rely on the definition of "representation" as "linearly readable information". 
%
We employ the encoding analysis procedure introduced by ~\citet{naselaris2011encoding} to evaluate the representational similarity between an AI model and brain recordings.
This linear model seeks to find whether there exists a linear mapping $W\in\mathbb{R}^{m\times d}$ that reliably predicts $m$-dimensional brain activity ($Y\in\mathbb{R}^{n \times m}$) given the $d$-dimensional model activation ($X\in\mathbb{R}^{n\times d}$) in response to $n$ images: 

\begin{equation*}
    \underset{W}{\arg\min} \left\{ \| Y - XW \|_2^2 + \lambda \| W \|_2^2 \right\}    
\end{equation*}
with $\lambda$ the ridge regularization parameter. 

For this, we use scikit-learn's \texttt{RidgeCV} \citep{pedregosa2011scikit} and 10 logarithmically-spaced regularization  $\lambda$ in between $10^0$ and $10^{8}$, and a 5-split cross-validation.

\subsection{Metrics}

\paragraph{Encoding score}
Given two representations $X$ and $Y$, we quantify their overall representational similarity by computing, for each split separately and then averaged, an \emph{encoding score} with a Pearson correlation score $R\in[-1, 1]$:
\begin{equation*}
    R^{(d)}=corr(WX_{test}, y_{test}^{(d)})
\end{equation*}

For clarity, we can either summarize the average R score across brain dimensions, or plot them all separately to get information about where brain activations are linearly predictable from the model. 
In some analysis, we use $\tilde{R}=R/max(R)$, the normalized encoding score, which peaks at 1.


\paragraph{Spatial score}
To assess whether a model organized its processing hierarchy similar to that of a brain with a \emph{spatial score}, we proceed in four steps.
%
First, we evaluate an encoding score for each dimension $d$ of the brain, and from 22 layers $k \in [0, 1]$ of the model, where 0 is the first layer, and 1 is the last layer.
Second, we identify the layer that best predicts this brain response: $k^*$.
Third, we approximate the hierarchical position $d^*$ of each brain region, as its Euclidean distance from V1 in the standardized MNI space, in mm. Note that this is a coarse approximation, as the actual cortical hierarchy does not strictly follow such distances, and may be considerably more complex \citep{felleman1991distributed}.
Finally, we compute the spatial score as the correlation between $d^*$ and $k^*$. For clarity we restrict these analyses to regions of interest. 

\paragraph{Temporal score}
To evaluate an analogous metrics from MEG recordings, we estimate a \emph{temporal score}: i.e. the correlation between the model layers $k$ and $t^*$ -- the time at which each layer of the model is maximally predictive of brain activity. To limit noisy estimate, we average of the temporal window during which \( \tilde{R}^{k} \geq 95\% \) where  \( \tilde{R}^{k}\) is the normalized brain-score of the layer $k$.

\subsection{Models}
\paragraph{Architecture.}
DINOv3 is a state-of-the-art self supervised learning vision transformer model trained on 1.7 billion natural images \citep{dinov3}. We train, from scratch, a selection of eight variants of this DINOv3 model to ensure a comprehensive evaluation ranging through architectures, training scale and data types.

First, we leverage the DINOv3-7B, trained across $1e^7$ checkpoints. We analyze comparatively DINO Small, Base, Large and Giant, after training for $5e^6$ training steps on 1.7B images with the same configuration.
Additionally, we train and analyze comparatively 3 versions of the DINO Large architecture: DINO Human, DINO Cellular and DINO Satellite. These models were configured similarly and trained, from scratch, over $5e^6$ steps on 10M images; they only differ in the type of images with which they were trained. 

\begin{table*}[htb]
\centering
\begin{tabular}{lcccc}
\hline
\textbf{Model} & \textbf{Parameters} & \textbf{Layers} & \textbf{Batch Size} & \textbf{Images}\\
\hline
DINOv3            & 7B    & 40 & 4096 &  Human centric 1.7B\\
DINOv3 Giant      & 1.1B  & 32 & 4096 &  Human centric 1.7B\\
DINOv3 Large      & 300M  & 24 & 4096 &  Human centric 1.7B\\
DINOv3 Base       & 86M   & 12 & 4096 &  Human centric 1.7B\\
DINOv3 Small      & 21M   & 12 & 4096 &  Human centric 1.7B\\
DINOv3 Human      & 300M  & 24 & 2048 &  Human centric 10M\\
DINOv3 Cellular   & 300M  & 24 & 2048 &  Cellular 10M\\
DINOv3 Satellite  & 300M  & 24 & 2048 &  Satellite 10M\\
\hline
\end{tabular}
\caption{Specifications of DINOv3 model variants.}
\label{tab:dino_models}
\end{table*}

\subsection{Datasets.}

\paragraph{Images.}
DINOv3-7B and Dino Human were trained on the same human-centric data. This dataset was constructed from a large pool of web images obtained from public Instagram posts, street views and ImageNet \citep{Deng2009}. These images went through platform-level content moderation to prevent harmful contents, in order to obtain an data pool of approximately 17 billion images. This data pool was curated following the procedure in \citep{dinov3} to obtain a large-scale pre-training dataset of 1.7 billion images. To compare models trained with different types of images, we re-trained three distinct large DINOv3 with one of three types of natural images -- human-centric, cellular and satellite images -- matched in terms of quantity (10M images each). 

Human-centric images correspond to the dataset used for training the original DINOv3 model. For our comparative analyses on human-centric, cellular and satellite images, we randomly selected from this dataset of 1.7 billion images a subset of 10 million images.

Cellular images correspond to the ExtendedCHAMMI dataset, which consists of fluorescent microscopic images of cells revealing cellular structures into different channels (e.g. nucleus, mitochondria, microtubules, etc.) \citep{cellular_dino}. 

Satellite images correspond to a random subset of the SAT-493M dataset, which consists of approximately 500 million sampled randomly from Maxar RGB ortho-rectified imagery at 0.6 meter resolution \citep{dinov3}.

\paragraph{Magnetoencephalography (MEG).} We use the THINGS-MEG dataset \citep{contier2023}, which consists of MEG recordings from four healthy participants viewing 22,500 naturalistic images, representing a total of 1,800 object concepts \citep{hebart2023things}. Images were presented during 1.5\,s, while participants maintained fixation. To limit the impact of noise we apply a bandpass filter between 0.1 and 20\,Hz, down-sample the signal at 30\,Hz, time-lock the brain responses to individual words, and epoch the corresponding neural data between -0.5\,s and +3\,s relative to word onset using MNE-Python \citep{gramfort2013meg}. Finally, we z-score MEG signals across words, for each MEG channel and each time point independently.\\
\emph{time ROIs}. We study individually three 5s-long time ROIs across the processing time of an image, to study the relative impact of each layer in the encoding of the cognitive process at play during that time. These time windows span .08-.13s, .13-.18s and .5-.55s.\\
\emph{$T^{layer}_{\text{max}}$}. To study the dynamics of each layer, we compute $T^{layer}_{\text{max}}$, the mean of the temporal window during which \( \tilde{R}^{layer} \geq 95\% \) where  \( \tilde{R}^{layer}\) is the normalized brain-score of each layer.

\paragraph{Functional Magnetic Resonance Imaging (fMRI).} We leverage the Natural Scenes Dataset \citep{Allen2022}, a 7 tesla  fMRI dataset which consists of recordings from eight subjects, each observing a total of 10 000 natural scenes during 4 seconds each, while performing a continuous recognition task. We encode the BOLD signal on the fsaverage surface at 5.5\,s after image onset. This timestep corresponds to the peak of decoding of the image from the BOLD signal.

\emph{Regions of interest (ROIs)}. For clarity, we select a representative set of 15 regions of interest (ROIs) spanning the anatomy of the cortex, among the regions encoded with an averaged FDR-corrected t-test p < 0.01, among voxels forming the ROI). These ROIs are distributed from posterior-occipital lobe to prefrontal cortex. 
To investigate the cortical properties that index representational similarity, 
we analyze our results in light of four cortical maps, made available through Neuromaps \citep{markello2022neuromaps}: \\
\emph{Cortical expansion} \citep{hill2010similar} reflects the difference of cortical surface area between infants and adults. \\
\emph{Myelin concentration} is estimated from the T1w/T2w ratio in the HCP S1200 dataset \citep{van2013wu}. \\
\emph{Intrinsic timescales} are derived from mapping electromagnetic networks (measured through MEG) to hemodynamic network (measured through fMRI) and indexing the temporal integration window of each region \citep{Shafiei2021}. \\
\emph{Cortical thickness} is estimated by measuring the distance between "white" and "pial" Freesurfer \citep{freesurfer} surfaces from structural MRI in the Human Connectome Project \citep{van2013wu}.

\subsection{Statistics}
\emph{fMRI voxels.} We only plot and analyze voxels thresholded with p < 0.01 after a FDR-corrected t-test.\\
\emph{Across subjects.} To evaluate statistical estimates across subjects, we perform a Wilcoxon test using scipy \citep{virtanen2020scipy}. To correct for multiple comparison, we apply a false discovery rate correction, as implemented in MNE-Python \citep{gramfort2013meg}.  \\
\emph{Half times.} To analyze the speed of convergence of DINO models during training, we estimate the `half time`: the relative training step at which the similarity metric reaches half of its final value.

\section{Results}

\begin{figure*}[htb]
    \centering
    \includegraphics[width=1.0\linewidth]{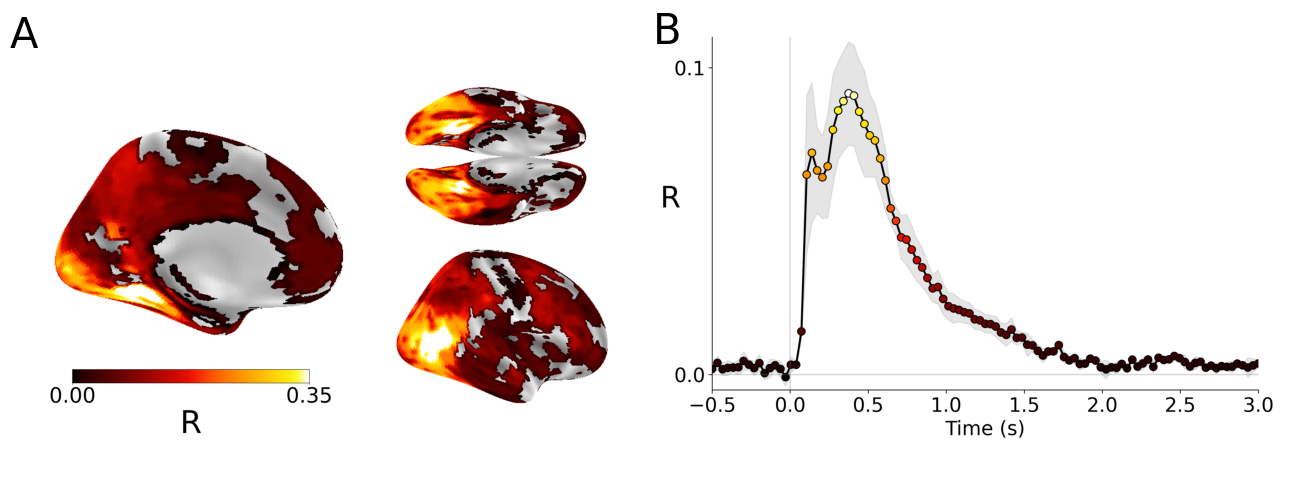}
    \caption{
    \textbf{Brain-DINOv3 similarity.}
    \textbf{A.} Similarity between DINOv3 embedding and the fMRI responses to corresponding images as estimated with a Pearson Brain-Score, and  fdr-corrected-thresholded at p < 0.01 (left: medial view of left hemisphere, top right: bottom view; bottom right: lateral view of right hemisphere). 
    \textbf{B.} Similarity between DINOv3 embedding and MEG responses to the corresponding images. The error bar indicates the standard error of the mean across 4 subjects watching still images.}
    \label{fig:encoding_score}
\end{figure*}

\subsection{DINOv3-brain similarity}

\paragraph{Encoding score.}
To verify that DINOv3 generates representations of natural images that are similar to those of the brain, we perform a cross-validated encoding analysis by evaluating the linear mapping between the activations of DINOv3 and of the brain in response to the same images.
%
Functional MRI results show that DINOv3 has representations that primarily peak in the visual pathway (R=$.45\pm.039$ - SEM across subjects), mostly in the lateral-occipitotemporal (MT: R=$.34\pm.026$) and ventromedial visual cortex (VMV2: R=$.28\pm.025$), Fig \ref{fig:encoding_score}A. 

%
MEG results show that this similarity rises around 70 ms after image onset (R=$.09\pm.017$, Fig \ref{fig:encoding_score}B) and remains significantly above chance level up to 3 seconds after image onset (p < 1e{-3}).

These results are consistent with past studies \citep{eickenberg2017seeing, schrimpf2018, Tang2025} and additionally show that areas typically discarded from the visual pathways, e.g. prefrontal regions BA 44, BA 45, IFSa and IFSp, also present activations that are linearly predictable from the AI embedding.

\begin{figure}[!htb]
    \centering
    \includegraphics[width=1.\columnwidth]{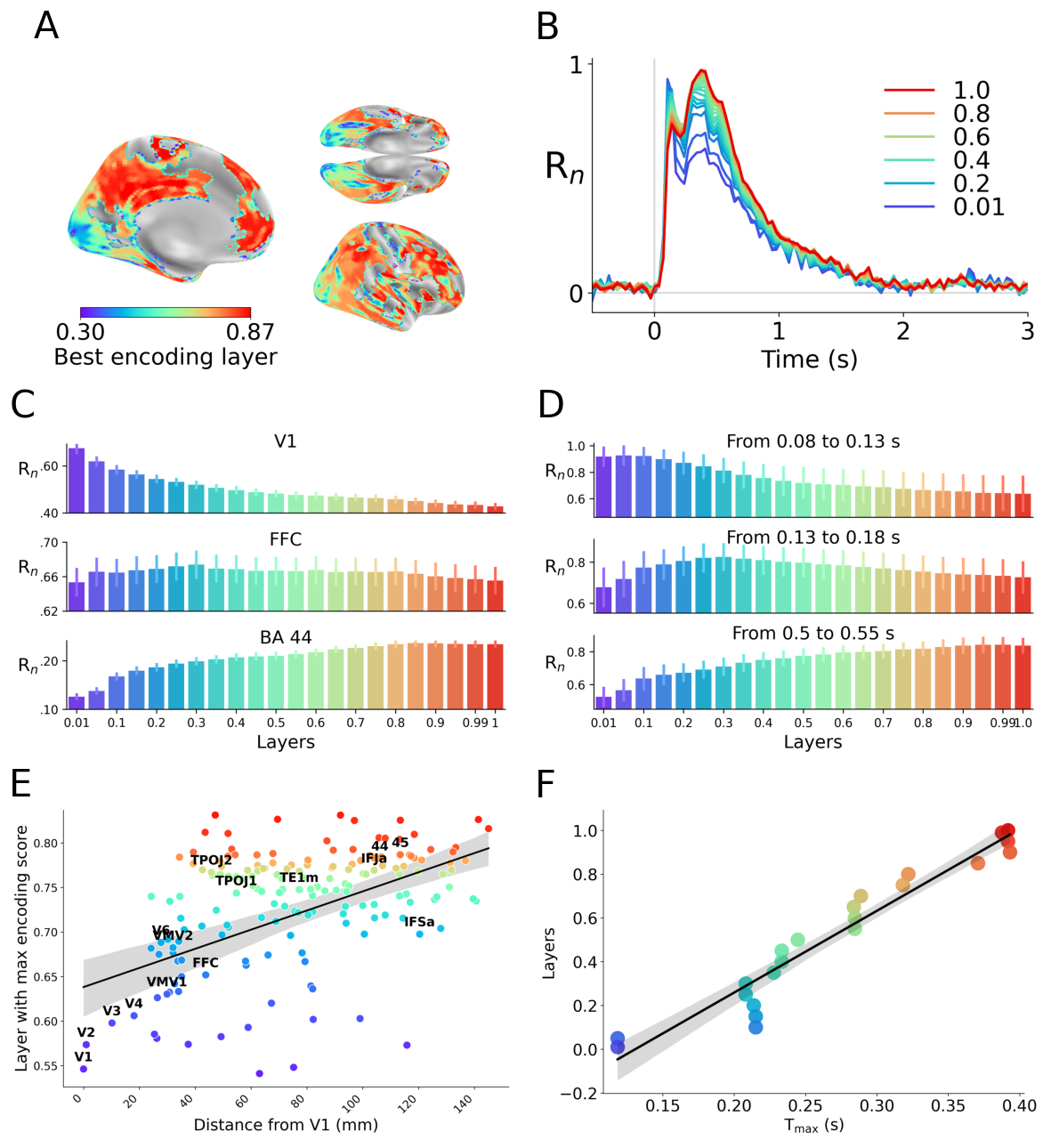}
    \caption{
    \textbf{The representational hierarchy of DINOv3 corresponds to the brain’s}
    \textbf{A.} Voxel-wise best encoding layers of DINOv3, FDR-corrected and thresholded at p < 0.01 (left: medial view of left hemisphere, top right: bottom view; bottom right: lateral view of right hemisphere). 
    \textbf{B.} Dynamic brain-score across time between each layer of DINOv3 and MEG responses to the corresponding still images.
    \textbf{C.} Scaled brain-scores across layers for three regions of interest across the cortex: V1 (up), Fusiform Face Cortex (middle) and Broadmann 44 (down). Encoding scores scores are scaled for each layer, by the maximal encoding score of this layer among the 15 studied ROIs. 
    \textbf{D.} Scaled brain-scores across layers for three time windows of interest during the processing of an image: .08-.12s (up), .13-.18s (middle) and .5-.55s (down). Encoding scores are scaled for each layer, by the maximal encoding score of this layer across time. 
    \textbf{E.} Plotting the correlation between the best encoding layer for each region and the euclidean distance of this region from V1, in mm. The  Pearson correlation is r = 0.38, p < 1e-6. Plotted regions are encoded with FDR-corrected thresholding at p < 0.01. 
    \textbf{F.} Plotting the correlation between the best encoding layer for each timestep and the $T^{layer}_{\text{max}}$, in s. The Pearson correlation is r = 0.84, p < 1e-5. Plotted regions are encoded with FDR-corrected thresholding at p < 0.01. }
    \label{fig:hierarchy}
\end{figure}

\paragraph{Spatial score.}
Does the hierarchy of representations of DINOv3 correspond to the visual hierarchy in the human brain? To address this question, we estimate the "spatial score".
%
The fMRI results confirm that the lowest layers of DINOv3 tend to best predict the lower-level sensory regions such as V1, whereas the highest layers tend to best predict higher-level regions of the brain, such as the prefrontal cortex (Fig \ref{fig:hierarchy}A, C, E).
A Pearson correlation between (i) the Euclidean distance between each brain region and V1, and (ii) the best encoding layer is highly significant, R=0.38, p < $1e^{-6}$ (Fig \ref{fig:hierarchy}E).

\paragraph{Temporal score.}
To complement this fMRI "spatial score", we evaluate an MEG "temporal score". For this, we identify the layer which best predicts each time ROIs relative to image onset in the MEG (Fig \ref{fig:hierarchy}B). The results show a significant correlation between $T^{layer}_{\text{max}}$ and layers, hereafter referred to as the temporal score (Fig \ref{fig:hierarchy}B, D, F). The temporal score R=0.96, p< $1e^{-12}$, shows that the first and last layers of DINOv3 consistently align with the earliest and latest MEG responses, respectively.

\subsection{What factors lead DINOv3 to become brain-like?}
\begin{figure*} 
    \centering
    \includegraphics[width=1.0\linewidth]{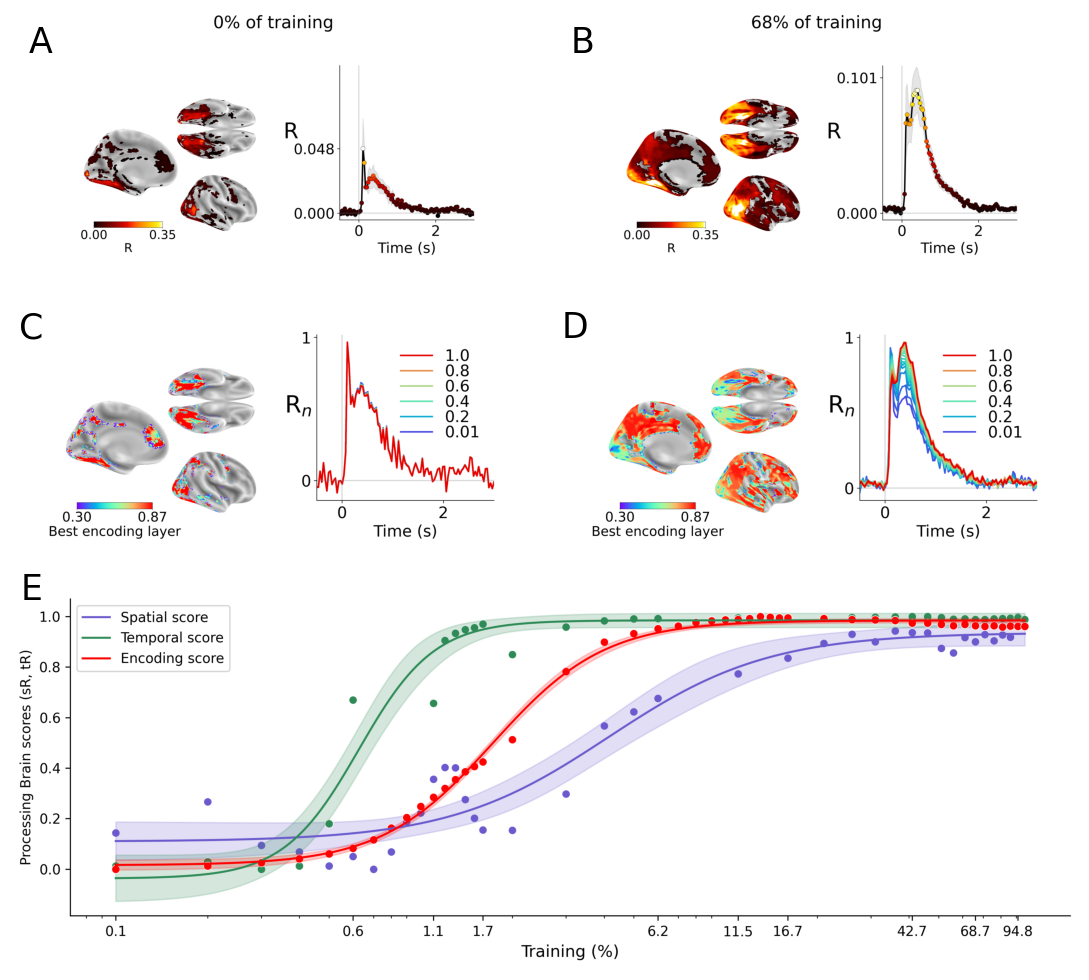}
    \caption{
    \textbf{A.} fMRI and MEG encoding scores from an untrained DINOv3.
    \textbf{B.} Encoding scores for a DINOv3 trained at 68\% of its training.
    \textbf{C.} Scores for an untrained DINOv3. Left. Best encoding layer for each voxel of each fMRI ROI. Right. Relative encoding score for 10 representative layers of an untrained DINOv3 (0=first layer, 1=last layer).
    \textbf{D.} Same as C for a DINOv3 trained at 68\% of its training. 
    \textbf{E.} Evolution of temporal, encoding and spatial scores as a function of DINOv3's training. 
    }
    \label{fig:training}
\end{figure*}
\paragraph{Impact of training.}
To clarify the emergence of brain-like processes in DINOv3, we evaluate the encoding score, spatial score and temporal score at each selected training step of DINOv3, and then summarize their developmental speed with a “half time”: i.e. the training step where half of the final score is reached.

\begin{figure*}
    \includegraphics[width=1.0\linewidth]{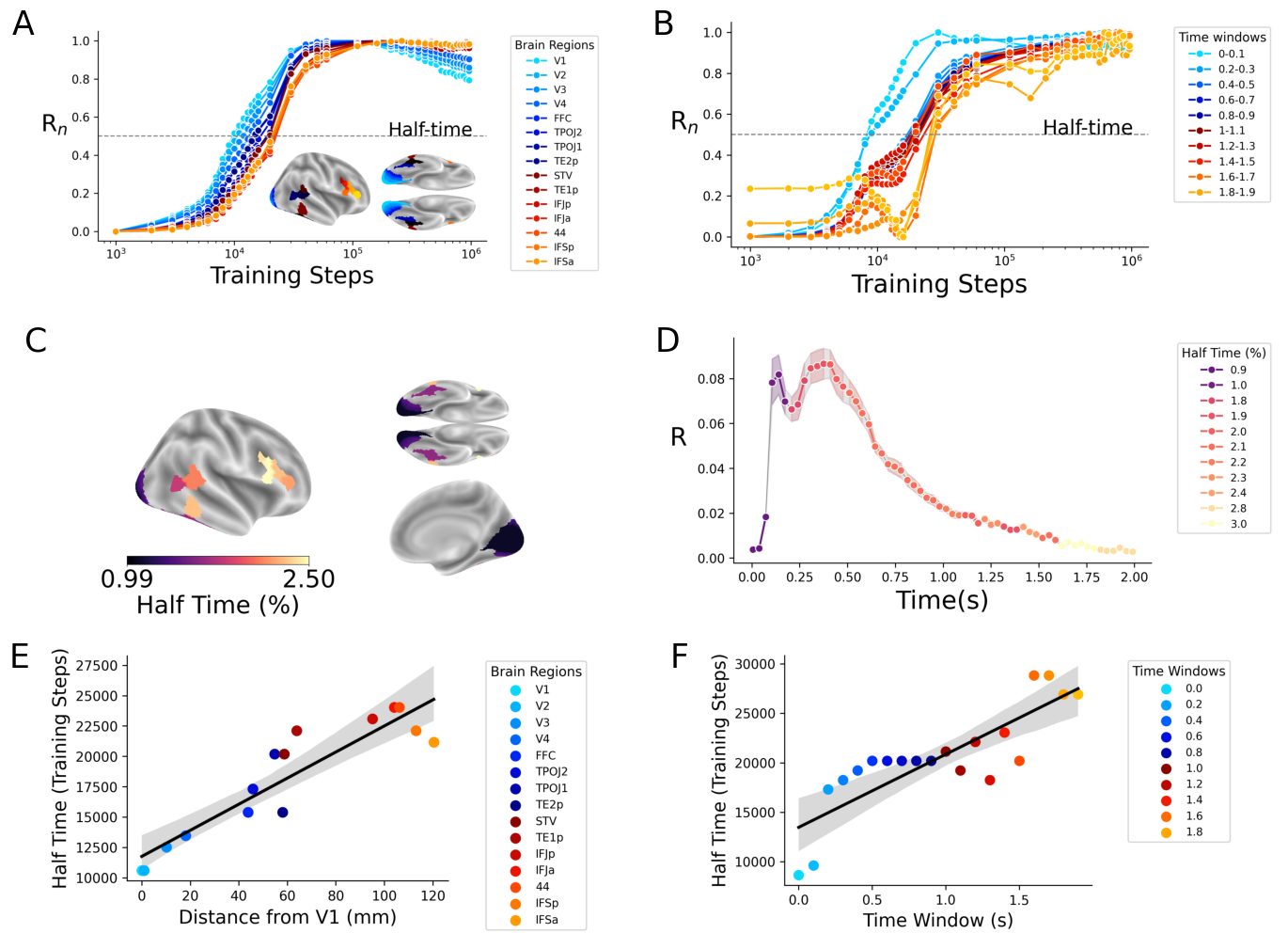}
    \caption{
    Emergence of brain-like representations. 
    \textbf{A.} Normalized brain encoding scores as a function of training for each brain ROI. The dashed line indicates the 50\% of the maximum encoding score for each region. 
    \textbf{B.} Same as A for MEG time region of interest.
    \textbf{C.} Half time for each brain region of interest.
    \textbf{D.} Half time for each time region of interest. 
    \textbf{E.} Correlation between half time of encoding score across training, and distance of each ROI from V1.
    \textbf{F.} Correlation between half time of encoding score across training, and time position of the encoded cognitive process.
    }
    
    \label{fig:halftime}
\end{figure*}

First, before training the encoding score reaches R=$.03 \pm 2e^{-4}$, after training it ultimately converges to R=$.09\pm5e^{-4}$. These R-scores are averaged across voxels – the best voxel peaking at R=$.45\pm.038$(Fig \ref{fig:training}A, B, E). The half time of the encoding score occurs around 2\% of the training, around $10^5$ training steps (i.e. ~800 million images).
Second, the temporal scores emerge faster than brain-scores: 
with a half time around 0.7\% of the training, and a convergence at R=0.96  (p$<1e^{-12}$).
Finally, the spatial score 
reaches its half time at 4\% of the training and converges to R=0.4 (p$<1e^{-6}$).
%
Are these developmental trajectories identical across temporal and brain regions of interest ? To address this issue, we evaluate the same analyses on specific regions or temporal windows of interest.
Functional MRI results show that low-level visual regions (e.g. V1, V2) are marked by lower half times than high-level prefrontal cortices (e.g. IFSp, IFSa; Fig \ref{fig:halftime}A,C). The correlation between half time and anatomical location (coarsely defined as the Euclidean distance to V1) is R=0.91, p$<1e^{-5}$.  
Similarly for MEG, earlier windows (e.g. $<$200\,ms) are marked by lower half times than late time windows (e.g. $>$1,500\,ms; Fig \ref{fig:halftime}B,D). The correlation between half time and temporal peak is R=0.84, p$<1e^{-5}$. 
Overall, these results show that the brain responses of the sensory and prefrontal cortices contain representations of images that are acquired relatively early and late in the training of DINOv3, respectively.

\paragraph{Impact of model size.}
 \begin{figure*}
    \centering
    \includegraphics[width=0.9\linewidth]{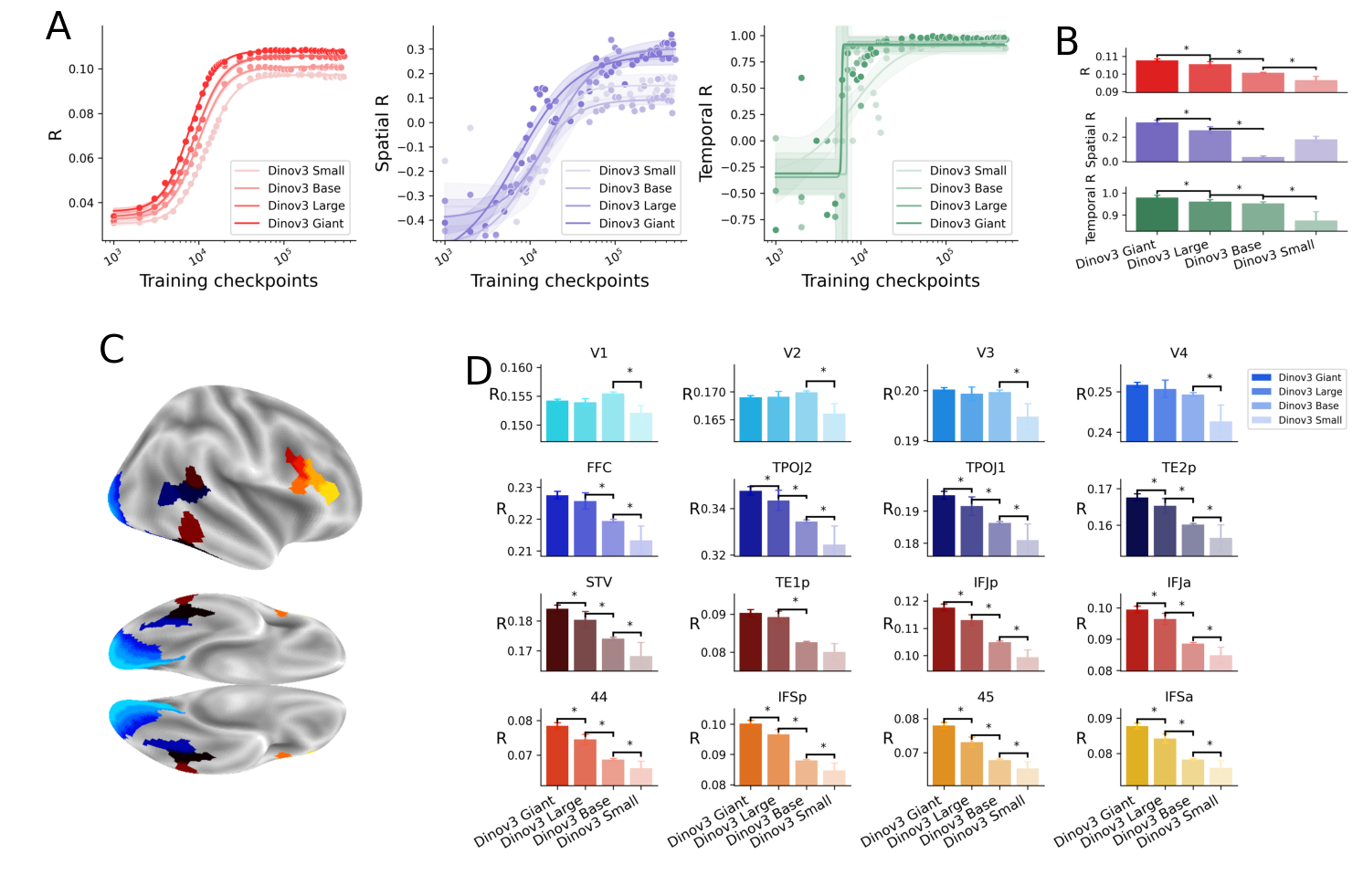}
    \caption{Impact of model size. For inter-model comparisons, significance to p<$1e^{-3}$ are represented by asterisks $*$.
    \textbf{A.} Encoding (reds), spatial (purples), and temporal scores (greens) as a function of training and model size. Logarithmic fits of scores across training.
    \textbf{B.} Scores on the final k=$4e^{5}$ training steps. 
    \textbf{C.} Brain ROIs.
    \textbf{D.} Encoding scores for each ROI at the end of training. 
    }
    \label{fig:modelsize}
\end{figure*}
Dino models of larger scale appear to converge quicker and encode higher-level ROIs more accurately. 
How does model size impact convergence? 
Model size consistently leads to bigger encoding scores at the end of training ($R_{\mathrm{Giant}} = 0.107 
> R_{\mathrm{Large}} = 0.105 
> R_{\mathrm{Base}} = 0.101 
> R_{\mathrm{Small}} = 0.096$ with p < $1e^{-3}$), Fig \ref{fig:modelsize}B). 
Similar, although noisier phenomena can be observed for spatial scores and temporal scores (p < $1e^{-3}$), Fig \ref{fig:modelsize}A, B.

Does model size impact encoding scores similarly across ROIs? Applying the same analysis for each ROI separately (Fig \ref{fig:modelsize}C, D) shows that model size primarily increases encoding of higher-level cortices like  BA44 and IFS as compared to visual cortices like V1, V2. All models present this size-dependent increased encoding significantly in higher-ROIs, only the smallest ones in V1, V2 (p < $1e^{-3}$), Fig \ref{fig:modelsize}C, D.

\paragraph{Impact of image type.}
\begin{figure*}
    \centering
    \includegraphics[width=.9\linewidth]{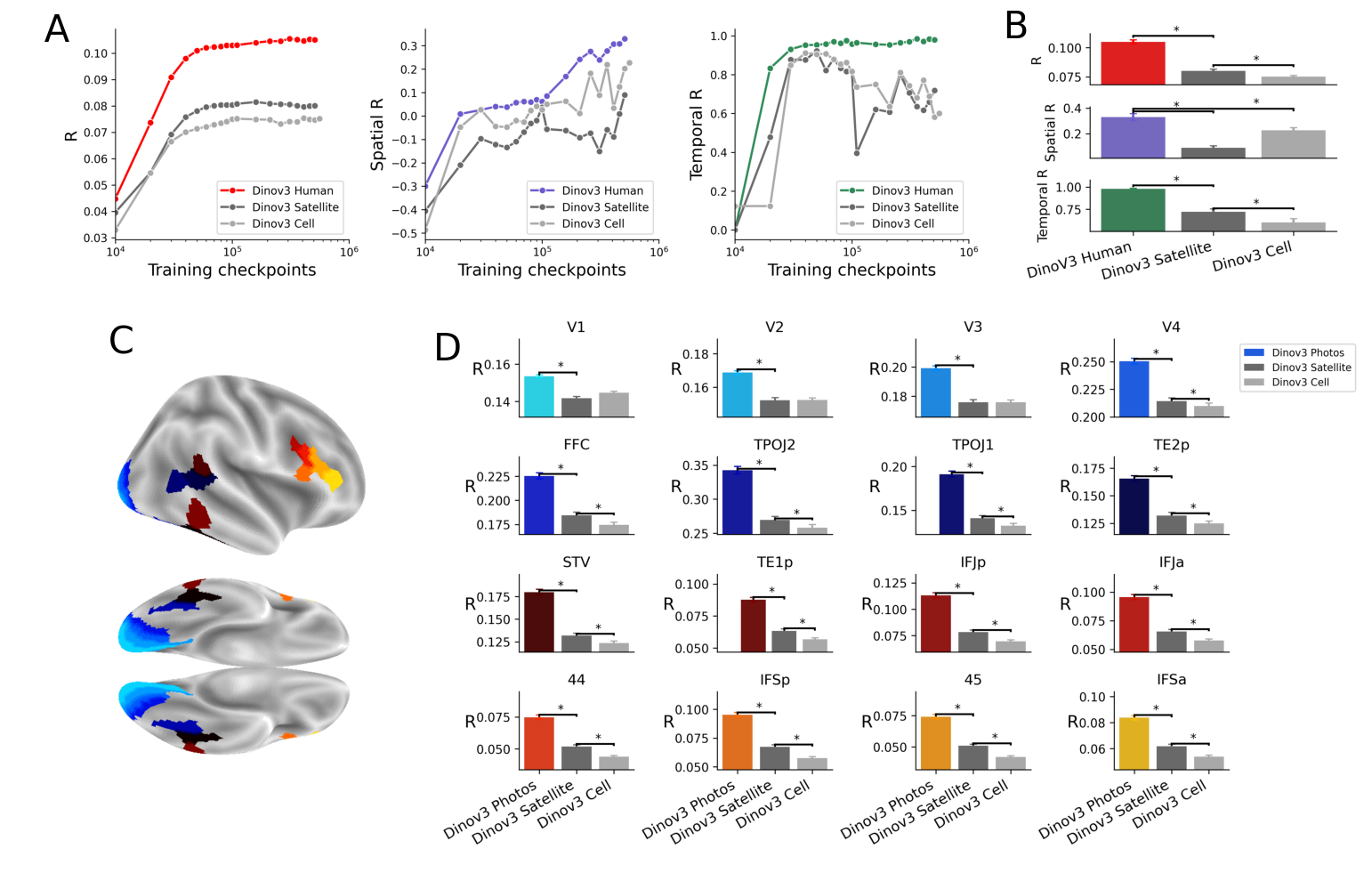}
    \caption{Impact of image type. For inter-model comparisons, significance to p<$1e^{-3}$ are represented by asterisks $*$.
    \textbf{A.} Encoding (reds), spatial (purples), and temporal scores (greens) as a function of training and image type.
    \textbf{B.} Scores on the final k=$4e^{5}$ training steps.
    \textbf{C.} Brain ROIs.
    \textbf{D.} Encoding scores for each ROI at the end of training.
    }
    
    \label{fig:image_type}
\end{figure*}
To assess how image types influence the development of brain-like representations in a model, we trained, from scratch, three distinct DINO models, each using one of three natural images datasets: satellite images, cell images and classic (human-centric) images. 
We focus on a single DINOv3 architecture (Dino Large), with a fixed training length and training data quantity (10M images) and have data type as the only varying factor.
Training improves encoding scores, spatial scores and temporal scores for all image types (Fig \ref{fig:image_type}A), suggesting that these models learn visual features that are universal across these different types of natural images.
However, these brain-similarity metrics are lower for satellite and cell images than for human centric images, for encoding, spatial and temporal scores. 
Interestingly, this difference is observed across most regions of interest: e.g. both V1 (p < $1e^{-3}$) and IFSa (p < $1e^{-3}$) are better encoded by a model trained with human centric photos than other models.
%
At the end of training, Dino Human reaches a significantly higher performance regarding brain-score, temporal and spatial scores (p < $1e^{-3}$), Fig \ref{fig:image_type}B. These results might unravel from the fact that human centric images reflect visual input that brains are exposed to, whereas satellite images and cell images are images that brains have not been trained to process. 


\subsection{Link to cortical properties}
Is the development of brain-like representations predicted by functional, structural and developmental properties of the cortex? To explore this issue, we evaluate the correlation between the representational half time of encoding and four properties of the cortex.

\begin{figure*}
    \centering
    \includegraphics[width=1.0\linewidth]{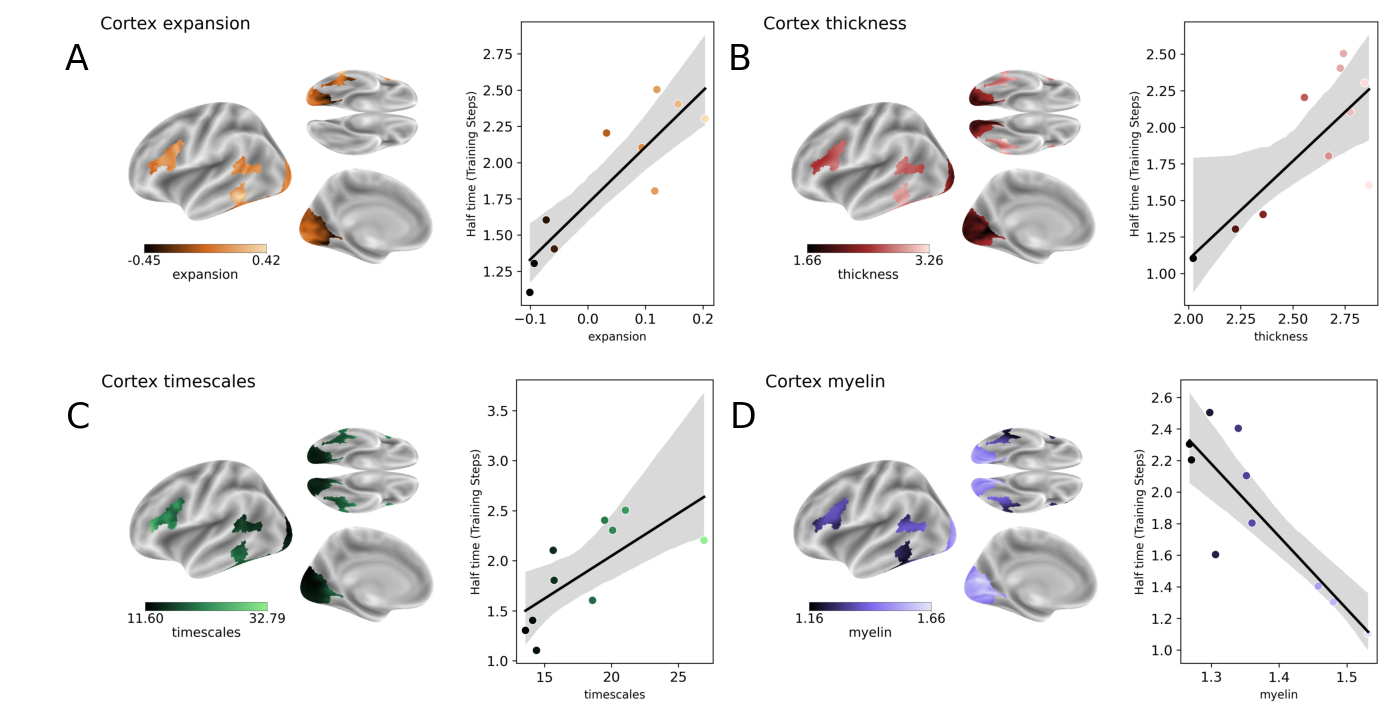}
    \caption{Relation between shared representations and cortical properties.
    \textbf{A.} Left. Cortical expansion index, as estimated from the difference between adults and infants' brains, for each ROI \citep{hill2010similar}. Right. Correlation between cortical expansion and half time. Each dot is an ROI.
    \textbf{B.} Same as A for cortical thickness, as estimated from \citep{van2013wu}.
    \textbf{C.} Same as A for cortical time scales, as estimated from MEG source reconstruction in \citep{Shafiei2021}.
    \textbf{D.} Same as A for myelin concentration, as estimated from \citep{van2013wu}.
    }
    \label{fig:cortex_properties}
\end{figure*}

\paragraph{Cortical expansion.} First, we focus on the developmental expansion of cortical regions. Using an atlas comparing infant and adult cortical structures \citep{hill2010similar}, we found a strong positive correlation (R=0.88, p < $1e^{-3}$) between half time and cortical expansion (Fig \ref{fig:cortex_properties}A). This indicates that cortical areas marked by greater developmental growth are also those whose representations emerge later in the AI model.

\paragraph{Cortical thickness.} Second, we assess the correspondence with cortical thickness, utilizing HCP S12000 estimates. Our results show a significant correlation (R=0.77, p < $1e^{-2}$), suggesting that cortical areas with larger cortical sheets exhibit longer half times (Fig \ref{fig:cortex_properties}B).

\paragraph{Cortical dynamics.} Third, the areas with the slowest intrinsic dynamics, as estimated from a source-reconstruction of MEG activity, are also those that tend to have the longest half times (R=0.71, p = .022). This result directly echoes our MEG results (Fig 5), whereby deeper layers of DINOv3 tend to be associated with slower brain responses (Fig \ref{fig:cortex_properties}C).

\paragraph{Cortical myelin.} Finally, this dynamic property appears linked to myelin concentration \citep{van2013wu}. Myelin, which facilitates faster neuronal transmission, demonstrated a strong negative correlation with half time (R=-0.85, p-val =$1e^{-3}$). This implies that higher myelin concentration is associated with shorter half times (Fig \ref{fig:cortex_properties}D).

In summary, these findings demonstrate a strong predictive relationship between the speed at which brain-like representations emerge in AI models and various structural and functional characteristics of the cortex, across development and once developed.

\section{Discussion}

\paragraph{Main findings.}
Understanding why artificial neural networks develop representations that resemble those in the human brain remains a fundamental challenge to neuroscience and AI \citep{platonic_hypothesis, Hasson2020, shen2025alignmentbrainsaievidence, caucheteux2022brains}. 
While recent studies have documented brain–model similarities across a wide range of architectures and training paradigms \citep{conwell2022can}, the exact factors that cause this convergence remain unclear. 
Here, we independently manipulate three independent factors – model size (from DINOv3 small to giant), training length (from 0 to $1e^7$ steps on several training sets of 10M and of 1.7B images) and image type (human-centric, satellite images and biological images) to test how each of them contributes to the emergence of brain-like representations of natural images. Our findings demonstrate that these three factors all independently and interactively impact the extent to which a self supervised model converges to brain-like visual processing.

\paragraph{Nativism and empiricism.}
In particular, the model–brain similarity increases consistently with larger architectures, longer training, and more ecologically valid data.
These results are consistent with an increasing set of studies showing linearly aligned representations of natural images \citep{yamins2014performance, Kriegeskorte2015, schrimpf2018, Tang2025, thobani2025modelbrain}, with a hierarchy that maps the functional organization of the visual cortices \citep{eickenberg2017seeing,la2022feature}, and dynamics that reflect the ordering of the model’s layers \citep{Seeliger2018, Cichy2016, caucheteux2022brains}.

In addition to its factorial disentanglement, our study provides additional contributions. \\
First, this model-brain alignment is not confined to the visual pathways \citep{eickenberg2017seeing, schrimpf2018, Tang2025} but extends into high-level -- multi-modal -- regions of the cortex, including the prefrontal cortex (although see e.g. \citet{solomon_semantic_2024} for a low-dimensional set of image features identified in the prefrontal lobe).\\
Second, our independent manipulation of model size, training duration, and data type further show how these factors \emph{interact} with one another: the largest architectures best align with brain activity as (1) they get trained and (2) on ecologically-relevant naturalistic images. \\ 
Third, even non-human-centric datasets (satellite images, biological images) support partial convergence in early visual areas, implying that low-level statistics shared across environments are sufficient to bootstrap early representations. 
Overall, these results suggest that while the architecture supplies a potential, the data remain critical in making these systems learn representations that are similar to the brain. This interaction between architectures, training and data provides an empirical framework to the  long-standing debates in cognitive science on nativism versus empiricism, – showing how ‘innate’ and ‘experiential’ interact with one another in the development of cognition.

\paragraph{Towards a model of the visual cortex ontogeny.}
This model-brain alignment follows a surprisingly steady developmental trajectory. Early in training, the models rapidly acquire representations that align with the fast and low-level visual responses of the sensory cortices. In contrast, the emergence of slow and high-level representations – particularly those aligning with the prefrontal cortex – appears to require both far more training data. \\
This developmental trajectory echoes the biological development of the human cortex: the brain areas with which the AI models align last during their training are precisely those with the greatest cortical thickness, slower intrinsic timescales, prolonged maturation, and lower levels of myelination – i.e. the areas of the associative cortices that are known to slowly develop throughout the first two decades of life \citep{dehaene2021we}. This result suggests that the sequential acquisition of representations in artificial neural networks may spontaneously model some of the developmental trajectories of brain functions. In doing so, they may ultimately provide a new computational framework to understand the staged maturation of visual processing in biological systems \citep{Vogelsang2024}. 

\paragraph{Open questions.}
Several results were not anticipated. First, the temporal score, encoding score and spatial score do not appear to emerge simultaneously – hence leading to the novel question of why these metrics follow this specific order. Second, the spatial and temporal scores are initially negative at the beginning of model training. This means that the deepest layers of a random DINOv3 tend to best predict fast and low-level brain responses at the very early (but not late) stages of training. Finally, the half times of these three metrics are reached in between 1\% and 4\% – i.e. only n=1.6B images – of DINOv3 training quantity. This suggests that while low-level brain-like representations are very quickly learnable, the high-level representations of the brain require a very large amount of data to be fully acquired.

\paragraph{Limitations.}
While this study offers a controlled analysis of brain–model convergence, several limitations warrant consideration. First, our findings are based exclusively on a single family of self-supervised vision models (DINOv3), which are hierarchical by design. It thus remains an open question whether similar spatial, temporal and encoding scores would emerge with other architectures and training objectives \citep{conwell2021what}. 
Second, fMRI and MEG offer limited resolution and thus provide coarse population-level brain activity and may overlook fine-grained neural mechanisms. 
Third, our analyses focus solely on the adult brain, leaving open the question of how these alignments emerge across development. Understanding when these correspondences arise will require data from infants, children, or longitudinal cohorts \citep{evanson2025emergence}. 
Finally, while we quantify the similarity between representations from models and the brain, the exact nature and semantic structure of these neuronal representations continues to be a subject of intense ongoing research \citep{Gifford2025, Graumann2022}.
Closing this interpretability gap certainly remains a major challenge to both neuroscience and AI.

\paragraph{Conclusion.}
Beyond the characterization of the spontaneous convergence between AI models and brains, these findings chart a path toward using AI models as tools to investigate the organizing principles of biological vision in the human brain. By showing how machines can come to see like us, our findings provide cues as to how the human brain may come to see the world.





\clearpage
\newpage
\bibliographystyle{assets/plainnat}
\bibliography{main}

\begin{thebibliography}{58}
\providecommand{\natexlab}[1]{#1}
\providecommand{\url}[1]{\texttt{#1}}
\expandafter\ifx\csname urlstyle\endcsname\relax
  \providecommand{\doi}[1]{doi: #1}\else
  \providecommand{\doi}{doi: \begingroup \urlstyle{rm}\Url}\fi

\bibitem[Allen et~al.(2022)Allen, St-Yves, Wu, Breedlove, Prince, Dowdle, Nau, Caron, Pestilli, Charest, Hutchinson, Naselaris, and Kay]{Allen2022}
Emily~J. Allen, Ghislain St-Yves, Yihan Wu, Jesse~L. Breedlove, Jacob~S. Prince, Logan~T. Dowdle, Matthias Nau, Brad Caron, Franco Pestilli, Ian Charest, J.~Benjamin Hutchinson, Thomas Naselaris, and Kendrick Kay.
\newblock A massive 7t fmri dataset to bridge cognitive neuroscience and artificial intelligence.
\newblock \emph{Nature Neuroscience}, 25\penalty0 (1):\penalty0 116–126, January 2022.
\newblock ISSN 1546-1726.
\newblock \doi{10.1038/s41593-021-00962-x}.
\newblock \url{http://dx.doi.org/10.1038/s41593-021-00962-x}.

\bibitem[Banville et~al.(2025)Banville, Benchetrit, d'Ascoli, Rapin, and King]{banville2025scaling}
Hubert Banville, Yohann Benchetrit, St{\'e}phane d'Ascoli, J{\'e}r{\'e}my Rapin, and Jean-R{\'e}mi King.
\newblock Scaling laws for decoding images from brain activity.
\newblock \emph{arXiv preprint arXiv:2501.15322}, 2025.

\bibitem[Cagnetta et~al.(2024)Cagnetta, Petrini, Tomasini, Favero, and Wyart]{Cagnetta2024}
Francesco Cagnetta, Leonardo Petrini, Umberto~M. Tomasini, Alessandro Favero, and Matthieu Wyart.
\newblock How deep neural networks learn compositional data: The random hierarchy model.
\newblock \emph{Physical Review X}, 14\penalty0 (3), July 2024.
\newblock ISSN 2160-3308.
\newblock \doi{10.1103/physrevx.14.031001}.
\newblock \url{http://dx.doi.org/10.1103/PhysRevX.14.031001}.

\bibitem[Caucheteux and King(2022)]{caucheteux2022brains}
Charlotte Caucheteux and Jean-R{\'e}mi King.
\newblock Brains and algorithms partially converge in natural language processing.
\newblock \emph{Communications biology}, 5\penalty0 (1):\penalty0 134, 2022.

\bibitem[Cheng et~al.(2022)Cheng, Misra, Schwing, Kirillov, and Girdhar]{cheng2021mask2former}
Bowen Cheng, Ishan Misra, Alexander~G. Schwing, Alexander Kirillov, and Rohit Girdhar.
\newblock Masked-attention mask transformer for universal image segmentation.
\newblock 2022.

\bibitem[Cichy et~al.(2016)Cichy, Khosla, Pantazis, Torralba, and Oliva]{Cichy2016}
Radoslaw~Martin Cichy, Aditya Khosla, Dimitrios Pantazis, Antonio Torralba, and Aude Oliva.
\newblock Comparison of deep neural networks to spatio-temporal cortical dynamics of human visual object recognition reveals hierarchical correspondence.
\newblock \emph{Scientific Reports}, 6\penalty0 (1), June 2016.
\newblock ISSN 2045-2322.
\newblock \doi{10.1038/srep27755}.
\newblock \url{http://dx.doi.org/10.1038/srep27755}.

\bibitem[Conwell et~al.(2021)Conwell, Prince, Alvarez, and Konkle]{conwell2021what}
Colin Conwell, Jacob~S. Prince, George~A. Alvarez, and Talia Konkle.
\newblock What can 5.17 billion regression fits tell us about artificial models of the human visual system?
\newblock In \emph{SVRHM 2021 Workshop @ NeurIPS}, 2021.
\newblock \url{https://openreview.net/forum?id=i_xiyGq6FNT}.

\bibitem[Conwell et~al.(2022)Conwell, Prince, Kay, Alvarez, and Konkle]{conwell2022can}
Colin Conwell, Jacob~S Prince, Kendrick~N Kay, George~A Alvarez, and Talia Konkle.
\newblock What can 1.8 billion regressions tell us about the pressures shaping high-level visual representation in brains and machines?
\newblock \emph{BioRxiv}, pages 2022--03, 2022.

\bibitem[Dehaene(2021)]{dehaene2021we}
Stanislas Dehaene.
\newblock \emph{How we learn: Why brains learn better than any machine... for now}.
\newblock Penguin, 2021.

\bibitem[Deng et~al.(2009)Deng, Dong, Socher, Li, Li, and Fei-Fei]{Deng2009}
Jia Deng, Wei Dong, Richard Socher, Li-Jia Li, Kai Li, and Li~Fei-Fei.
\newblock Imagenet: A large-scale hierarchical image database.
\newblock In \emph{2009 IEEE Conference on Computer Vision and Pattern Recognition}. IEEE, June 2009.
\newblock \doi{10.1109/cvpr.2009.5206848}.
\newblock \url{http://dx.doi.org/10.1109/CVPR.2009.5206848}.

\bibitem[DiCarlo et~al.(2012)DiCarlo, Zoccolan, and Rust]{dicarlo}
James~J DiCarlo, Davide Zoccolan, and Nicole~C Rust.
\newblock How does the brain solve visual object recognition?
\newblock \emph{Neuron}, 73\penalty0 (3):\penalty0 415--434, 2012.

\bibitem[Doerig et~al.(2025)Doerig, Kietzmann, Allen, Wu, Naselaris, Kay, and Charest]{Doerig2025}
Adrien Doerig, Tim~C. Kietzmann, Emily Allen, Yihan Wu, Thomas Naselaris, Kendrick Kay, and Ian Charest.
\newblock High-level visual representations in the human brain are aligned with large language models.
\newblock \emph{Nature Machine Intelligence}, August 2025.
\newblock ISSN 2522-5839.
\newblock \doi{10.1038/s42256-025-01072-0}.
\newblock \url{http://dx.doi.org/10.1038/s42256-025-01072-0}.

\bibitem[Eickenberg et~al.(2017)Eickenberg, Gramfort, Varoquaux, and Thirion]{eickenberg2017seeing}
Michael Eickenberg, Alexandre Gramfort, Ga{\"e}l Varoquaux, and Bertrand Thirion.
\newblock Seeing it all: Convolutional network layers map the function of the human visual system.
\newblock \emph{NeuroImage}, 152:\penalty0 184--194, 2017.

\bibitem[Esteva et~al.(2017)Esteva, Kuprel, Novoa, Ko, Swetter, Blau, and Thrun]{esteva2017dermatologist}
Andre Esteva, Brett Kuprel, Roberto~A Novoa, Justin Ko, Susan~M Swetter, Helen~M Blau, and Sebastian Thrun.
\newblock Dermatologist-level classification of skin cancer with deep neural networks.
\newblock \emph{nature}, 542\penalty0 (7639):\penalty0 115--118, 2017.

\bibitem[Evanson et~al.(2025)Evanson, Bulteau, Chipaux, Dorfmüller, Ferrand-Sorbets, Raffo, Rosenberg, Bourdillon, and King]{evanson2025emergence}
Linnea Evanson, Christine Bulteau, Mathilde Chipaux, Georg Dorfmüller, Sarah Ferrand-Sorbets, Emmanuel Raffo, Sarah Rosenberg, Pierre Bourdillon, and Jean-Rémi King.
\newblock Emergence of language in the developing brain.
\newblock Manuscript, May 2025.
\newblock \url{mailto:jeanremi@meta.com}.
\newblock Equal contribution by Pierre Bourdillon and Jean-Rémi King.

\bibitem[Felleman and Van~Essen(1991)]{felleman1991distributed}
Daniel~J Felleman and David~C Van~Essen.
\newblock Distributed hierarchical processing in the primate cerebral cortex.
\newblock \emph{Cerebral cortex (New York, NY: 1991)}, 1\penalty0 (1):\penalty0 1--47, 1991.

\bibitem[Fischl(2012)]{freesurfer}
Bruce Fischl.
\newblock Freesurfer.
\newblock \emph{NeuroImage}, 62\penalty0 (2):\penalty0 774–781, August 2012.
\newblock ISSN 1053-8119.
\newblock \doi{10.1016/j.neuroimage.2012.01.021}.
\newblock \url{http://dx.doi.org/10.1016/j.neuroimage.2012.01.021}.

\bibitem[Gifford et~al.(2025)Gifford, Jastrzębowska, Singer, and Cichy]{Gifford2025}
Alessandro~T. Gifford, Maya~A. Jastrzębowska, Johannes J.~D. Singer, and Radoslaw~M. Cichy.
\newblock In silico discovery of representational relationships across visual cortex.
\newblock \emph{Nature Human Behaviour}, June 2025.
\newblock ISSN 2397-3374.
\newblock \doi{10.1038/s41562-025-02252-z}.
\newblock \url{http://dx.doi.org/10.1038/s41562-025-02252-z}.

\bibitem[Gramfort et~al.(2013)Gramfort, Luessi, Larson, Engemann, Strohmeier, Brodbeck, Goj, Jas, Brooks, Parkkonen, et~al.]{gramfort2013meg}
Alexandre Gramfort, Martin Luessi, Eric Larson, Denis~A Engemann, Daniel Strohmeier, Christian Brodbeck, Roman Goj, Mainak Jas, Teon Brooks, Lauri Parkkonen, et~al.
\newblock Meg and eeg data analysis with mne-python.
\newblock \emph{Frontiers in Neuroinformatics}, 7:\penalty0 267, 2013.

\bibitem[Graumann et~al.(2022)Graumann, Ciuffi, Dwivedi, Roig, and Cichy]{Graumann2022}
Monika Graumann, Caterina Ciuffi, Kshitij Dwivedi, Gemma Roig, and Radoslaw~M. Cichy.
\newblock The spatiotemporal neural dynamics of object location representations in the human brain.
\newblock \emph{Nature Human Behaviour}, 6\penalty0 (6):\penalty0 796–811, February 2022.
\newblock ISSN 2397-3374.
\newblock \doi{10.1038/s41562-022-01302-0}.
\newblock \url{http://dx.doi.org/10.1038/s41562-022-01302-0}.

\bibitem[Hasson et~al.(2020)Hasson, Nastase, and Goldstein]{Hasson2020}
Uri Hasson, Samuel~A. Nastase, and Ariel Goldstein.
\newblock Direct fit to nature: An evolutionary perspective on biological and artificial neural networks.
\newblock \emph{Neuron}, 105\penalty0 (3):\penalty0 416–434, February 2020.
\newblock ISSN 0896-6273.
\newblock \doi{10.1016/j.neuron.2019.12.002}.
\newblock \url{http://dx.doi.org/10.1016/j.neuron.2019.12.002}.

\bibitem[Hebart et~al.(2023{\natexlab{a}})Hebart, Contier, Teichmann, Rockter, Zheng, Kidder, Corriveau, Vaziri-Pashkam, and Baker]{contier2023}
Martin~N. Hebart, Oliver Contier, Lina Teichmann, Adam~H. Rockter, Charles Zheng, Alexis Kidder, Anna Corriveau, Maryam Vaziri-Pashkam, and Chris~I. Baker.
\newblock "things-meg", 2023{\natexlab{a}}.

\bibitem[Hebart et~al.(2023{\natexlab{b}})Hebart, Contier, Teichmann, Rockter, Zheng, Kidder, Corriveau, Vaziri-Pashkam, and Baker]{hebart2023things}
Martin~N Hebart, Oliver Contier, Lina Teichmann, Adam~H Rockter, Charles~Y Zheng, Alexis Kidder, Anna Corriveau, Maryam Vaziri-Pashkam, and Chris~I Baker.
\newblock {THINGS}-data, a multimodal collection of large-scale datasets for investigating object representations in human brain and behavior.
\newblock \emph{eLife}, 12:\penalty0 e82580, feb 2023{\natexlab{b}}.
\newblock ISSN 2050-084X.
\newblock \doi{10.7554/eLife.82580}.
\newblock \url{https://doi.org/10.7554/eLife.82580}.

\bibitem[Hill et~al.(2010)Hill, Inder, Neil, Dierker, Harwell, and Van~Essen]{hill2010similar}
Jason Hill, Terrie Inder, Jeffrey Neil, Donna Dierker, John Harwell, and David Van~Essen.
\newblock Similar patterns of cortical expansion during human development and evolution.
\newblock \emph{Proceedings of the National Academy of Sciences}, 107\penalty0 (29):\penalty0 13135--13140, 2010.

\bibitem[Huh et~al.(2024)Huh, Cheung, Wang, and Isola]{platonic_hypothesis}
Minyoung Huh, Brian Cheung, Tongzhou Wang, and Phillip Isola.
\newblock The platonic representation hypothesis.
\newblock \emph{arXiv preprint arXiv:2405.07987}, 2024.

\bibitem[King and Dehaene(2014)]{king_dehaene}
Jean-R{\'e}mi King and Stanislas Dehaene.
\newblock Characterizing the dynamics of mental representations: the temporal generalization method.
\newblock \emph{Trends in cognitive sciences}, 18\penalty0 (4):\penalty0 203--210, 2014.

\bibitem[Kriegeskorte(2015)]{Kriegeskorte2015}
Nikolaus Kriegeskorte.
\newblock Deep neural networks: A new framework for modeling biological vision and brain information processing.
\newblock \emph{Annual Review of Vision Science}, 1\penalty0 (1):\penalty0 417–446, November 2015.
\newblock ISSN 2374-4650.
\newblock \doi{10.1146/annurev-vision-082114-035447}.
\newblock \url{http://dx.doi.org/10.1146/annurev-vision-082114-035447}.

\bibitem[Kriegeskorte et~al.(2008)Kriegeskorte, Mur, and Bandettini]{kriegeskorte}
Nikolaus Kriegeskorte, Marieke Mur, and Peter~A Bandettini.
\newblock Representational similarity analysis-connecting the branches of systems neuroscience.
\newblock \emph{Frontiers in systems neuroscience}, 2:\penalty0 249, 2008.

\bibitem[La~Tour et~al.(2022)La~Tour, Eickenberg, Nunez-Elizalde, and Gallant]{la2022feature}
Tom~Dupr{\'e} La~Tour, Michael Eickenberg, Anwar~O Nunez-Elizalde, and Jack~L Gallant.
\newblock Feature-space selection with banded ridge regression.
\newblock \emph{NeuroImage}, 264:\penalty0 119728, 2022.

\bibitem[Lorenci et~al.(2025)Lorenci, Yi, Moutakanni, Bojanowski, Couprie, Caicedo, and Pernice]{cellular_dino}
Alice V.~De Lorenci, Seung~Eun Yi, Th{\'e}o Moutakanni, Piotr Bojanowski, Camille Couprie, Juan~C. Caicedo, and Wolfgang Maximilian~Anton Pernice.
\newblock Scaling channel-adaptive self-supervised learning.
\newblock \emph{Transactions on Machine Learning Research}, 2025.
\newblock ISSN 2835-8856.
\newblock \url{https://openreview.net/forum?id=pT8sgtRVAf}.

\bibitem[Mahner et~al.(2025)Mahner, Muttenthaler, G\"{u}\c{c}l\"{u}, and Hebart]{Mahner2025}
Florian~P. Mahner, Lukas Muttenthaler, Umut G\"{u}\c{c}l\"{u}, and Martin~N. Hebart.
\newblock Dimensions underlying the representational alignment of deep neural networks with humans.
\newblock \emph{Nature Machine Intelligence}, 7\penalty0 (6):\penalty0 848–859, June 2025.
\newblock ISSN 2522-5839.
\newblock \doi{10.1038/s42256-025-01041-7}.
\newblock \url{http://dx.doi.org/10.1038/s42256-025-01041-7}.

\bibitem[Markello et~al.(2022)Markello, Hansen, Liu, Bazinet, Shafiei, Su{\'a}rez, Blostein, Seidlitz, Baillet, Satterthwaite, et~al.]{markello2022neuromaps}
Ross~D Markello, Justine~Y Hansen, Zhen-Qi Liu, Vincent Bazinet, Golia Shafiei, Laura~E Su{\'a}rez, Nadia Blostein, Jakob Seidlitz, Sylvain Baillet, Theodore~D Satterthwaite, et~al.
\newblock Neuromaps: structural and functional interpretation of brain maps.
\newblock \emph{Nature Methods}, 19\penalty0 (11):\penalty0 1472--1479, 2022.

\bibitem[Mehrer et~al.(2020)Mehrer, Spoerer, Kriegeskorte, and Kietzmann]{Mehrer2020}
Johannes Mehrer, Courtney~J. Spoerer, Nikolaus Kriegeskorte, and Tim~C. Kietzmann.
\newblock Individual differences among deep neural network models.
\newblock \emph{Nature Communications}, 11\penalty0 (1), November 2020.
\newblock ISSN 2041-1723.
\newblock \doi{10.1038/s41467-020-19632-w}.
\newblock \url{http://dx.doi.org/10.1038/s41467-020-19632-w}.

\bibitem[Millet et~al.(2023)Millet, Caucheteux, Orhan, Boubenec, Gramfort, Dunbar, Pallier, and King]{millet2023realisticmodelspeechprocessing}
Juliette Millet, Charlotte Caucheteux, Pierre Orhan, Yves Boubenec, Alexandre Gramfort, Ewan Dunbar, Christophe Pallier, and Jean-Remi King.
\newblock Toward a realistic model of speech processing in the brain with self-supervised learning, 2023.
\newblock \url{https://arxiv.org/abs/2206.01685}.

\bibitem[Naselaris et~al.(2011)Naselaris, Kay, Nishimoto, and Gallant]{naselaris2011encoding}
Thomas Naselaris, Kendrick~N Kay, Shinji Nishimoto, and Jack~L Gallant.
\newblock Encoding and decoding in {fMRI}.
\newblock \emph{Neuroimage}, 56\penalty0 (2):\penalty0 400--410, 2011.

\bibitem[Nikolaus et~al.(2024)Nikolaus, Mozafari, Asher, Reddy, and VanRullen]{nikolaus2024modalityagnosticfmridecodingvision}
Mitja Nikolaus, Milad Mozafari, Nicholas Asher, Leila Reddy, and Rufin VanRullen.
\newblock Modality-agnostic fmri decoding of vision and language, 2024.
\newblock \url{https://arxiv.org/abs/2403.11771}.

\bibitem[Pedregosa et~al.(2011)Pedregosa, Varoquaux, Gramfort, Michel, Thirion, Grisel, Blondel, Prettenhofer, Weiss, Dubourg, Vanderplas, Passos, Cournapeau, Brucher, Perrot, and Duchesnay]{pedregosa2011scikit}
F.~Pedregosa, G.~Varoquaux, A.~Gramfort, V.~Michel, B.~Thirion, O.~Grisel, M.~Blondel, P.~Prettenhofer, R.~Weiss, V.~Dubourg, J.~Vanderplas, A.~Passos, D.~Cournapeau, M.~Brucher, M.~Perrot, and E.~Duchesnay.
\newblock Scikit-learn: Machine learning in {P}ython.
\newblock \emph{Journal of Machine Learning Research}, 12:\penalty0 2825--2830, 2011.

\bibitem[Rajesh et~al.(2024)Rajesh, Jacob, and Arun]{rajesh2024brainlikeemergentpropertiesdeep}
Niranjan Rajesh, Georgin Jacob, and SP~Arun.
\newblock Brain-like emergent properties in deep networks: impact of network architecture, datasets and training, 2024.
\newblock \url{https://arxiv.org/abs/2411.16326}.

\bibitem[Redmon et~al.(2016)Redmon, Divvala, Girshick, and Farhadi]{redmon2016lookonceunifiedrealtime}
Joseph Redmon, Santosh Divvala, Ross Girshick, and Ali Farhadi.
\newblock You only look once: Unified, real-time object detection, 2016.
\newblock \url{https://arxiv.org/abs/1506.02640}.

\bibitem[Schrimpf et~al.(2018)Schrimpf, Kubilius, Hong, Majaj, Rajalingham, Issa, Kar, Bashivan, Prescott-Roy, Geiger, Schmidt, Yamins, and DiCarlo]{schrimpf2018}
Martin Schrimpf, Jonas Kubilius, Ha~Hong, Najib~J. Majaj, Rishi Rajalingham, Elias~B. Issa, Kohitij Kar, Pouya Bashivan, Jonathan Prescott-Roy, Franziska Geiger, Kailyn Schmidt, Daniel L.~K. Yamins, and James~J. DiCarlo.
\newblock Brain-score: Which artificial neural network for object recognition is most brain-like?
\newblock September 2018.
\newblock \doi{10.1101/407007}.
\newblock \url{http://dx.doi.org/10.1101/407007}.

\bibitem[Seeliger et~al.(2018)Seeliger, Fritsche, G\"{u}\c{c}l\"{u}, Schoenmakers, Schoffelen, Bosch, and van Gerven]{Seeliger2018}
K.~Seeliger, M.~Fritsche, U.~G\"{u}\c{c}l\"{u}, S.~Schoenmakers, J.-M. Schoffelen, S.E. Bosch, and M.A.J. van Gerven.
\newblock Convolutional neural network-based encoding and decoding of visual object recognition in space and time.
\newblock \emph{NeuroImage}, 180:\penalty0 253–266, October 2018.
\newblock ISSN 1053-8119.
\newblock \doi{10.1016/j.neuroimage.2017.07.018}.
\newblock \url{http://dx.doi.org/10.1016/j.neuroimage.2017.07.018}.

\bibitem[Shafiei et~al.(2021)Shafiei, Baillet, and Misic]{Shafiei2021}
Golia Shafiei, Sylvain Baillet, and Bratislav Misic.
\newblock Human electromagnetic and haemodynamic networks systematically converge in unimodal cortex and diverge in transmodal cortex.
\newblock September 2021.
\newblock \doi{10.1101/2021.09.07.458941}.
\newblock \url{http://dx.doi.org/10.1101/2021.09.07.458941}.

\bibitem[Shen et~al.(2025)Shen, Zhao, Dong, Zhang, and Zeng]{shen2025alignmentbrainsaievidence}
Guobin Shen, Dongcheng Zhao, Yiting Dong, Qian Zhang, and Yi~Zeng.
\newblock Alignment between brains and ai: Evidence for convergent evolution across modalities, scales and training trajectories, 2025.
\newblock \url{https://arxiv.org/abs/2507.01966}.

\bibitem[Sim{\'e}oni et~al.(2025)Sim{\'e}oni, Vo, Seitzer, Baldassarre, Oquab, Jose, Khalidov, Szafraniec, Yi, Ramamonjisoa, Francisco, Haziza, Wehrstedt, Wang, Darcet, Moutakanni, Sentana, Roberts, Vedaldi, Tolan, Brandt, Couprie, Mairal, J{\'e}gou, Labatut, and Bojanowski]{dinov3}
Oriane Sim{\'e}oni, Huy~V. Vo, Maximilian Seitzer, Federico Baldassarre, Maxime Oquab, Cijo Jose, Vasil Khalidov, Marc Szafraniec, Seungeun Yi, Michaël Ramamonjisoa, Massa Francisco, Daniel Haziza, Luca Wehrstedt, Jianyuan Wang, Timothée Darcet, Théo Moutakanni, Leonel Sentana, Claire Roberts, Andrea Vedaldi, Jamie Tolan, John Brandt, Camille Couprie, Julien Mairal, Herv{\'e} J{\'e}gou, Patrick Labatut, and Piotr Bojanowski.
\newblock Dinov3.
\newblock \emph{arXiv preprint arXiv:2025.00000}, 2025.
\newblock TL;DR: DINOv3.

\bibitem[Simkova et~al.(2025)Simkova, Doerig, Hickey, and Charest]{simkova2025representationsvisionlanguageconverge}
Katerina~Marie Simkova, Adrien Doerig, Clayton Hickey, and Ian Charest.
\newblock Representations in vision and language converge in a shared, multidimensional space of perceived similarities, 2025.
\newblock \url{https://arxiv.org/abs/2507.21871}.

\bibitem[Solomon et~al.(2024)Solomon, Kay, and Schapiro]{solomon_semantic_2024}
S.H. Solomon, K.~Kay, and A.C. Schapiro.
\newblock Semantic plasticity across timescales in the human brain.
\newblock \emph{bioRxiv}, 2024.
\newblock \doi{10.1101/2024.02.07.579310}.
\newblock \url{https://www.biorxiv.org/content/early/2024/05/24/2024.02.07.579310}.
\newblock Publisher: Cold Spring Harbor Laboratory.

\bibitem[Tang et~al.(2023)Tang, Du, Vo, Lal, and Huth]{tang2023brainencodingmodelsbased}
Jerry Tang, Meng Du, Vy~A. Vo, Vasudev Lal, and Alexander~G. Huth.
\newblock Brain encoding models based on multimodal transformers can transfer across language and vision, 2023.
\newblock \url{https://arxiv.org/abs/2305.12248}.

\bibitem[Tang et~al.(2025)Tang, Gokce, Al-Karkari, Yamins, and Schrimpf]{Tang2025}
Yingtian Tang, Abdulkadir Gokce, Khaled~Jedoui Al-Karkari, Daniel Yamins, and Martin Schrimpf.
\newblock Many-two-one: Diverse representations across visual pathways emerge from a single objective.
\newblock July 2025.
\newblock \doi{10.1101/2025.07.22.664908}.

\bibitem[Thobani et~al.(2025)Thobani, Sagastuy-Brena, Nayebi, Prince, Cao, and Yamins]{thobani2025modelbrain}
Imran Thobani, Javier Sagastuy-Brena, Aran Nayebi, Jacob~S. Prince, Rosa Cao, and Daniel~LK Yamins.
\newblock Model-brain comparison using inter-animal transforms.
\newblock In \emph{8th Annual Conference on Cognitive Computational Neuroscience}, 2025.
\newblock \url{https://openreview.net/forum?id=bra729zCMm}.

\bibitem[Tschannen et~al.(2025)Tschannen, Gritsenko, Wang, Naeem, Alabdulmohsin, Parthasarathy, Evans, Beyer, Xia, Mustafa, Hénaff, Harmsen, Steiner, and Zhai]{tschannen2025siglip2multilingualvisionlanguage}
Michael Tschannen, Alexey Gritsenko, Xiao Wang, Muhammad~Ferjad Naeem, Ibrahim Alabdulmohsin, Nikhil Parthasarathy, Talfan Evans, Lucas Beyer, Ye~Xia, Basil Mustafa, Olivier Hénaff, Jeremiah Harmsen, Andreas Steiner, and Xiaohua Zhai.
\newblock Siglip 2: Multilingual vision-language encoders with improved semantic understanding, localization, and dense features, 2025.
\newblock \url{https://arxiv.org/abs/2502.14786}.

\bibitem[Van~Essen et~al.(2013)Van~Essen, Smith, Barch, Behrens, Yacoub, Ugurbil, Consortium, et~al.]{van2013wu}
David~C Van~Essen, Stephen~M Smith, Deanna~M Barch, Timothy~EJ Behrens, Essa Yacoub, Kamil Ugurbil, Wu-Minn~HCP Consortium, et~al.
\newblock The wu-minn human connectome project: an overview.
\newblock \emph{Neuroimage}, 80:\penalty0 62--79, 2013.

\bibitem[van Rossem and Saxe(2024)]{vanrossem2024representationsalignuniversalityrepresentation}
Loek van Rossem and Andrew~M. Saxe.
\newblock When representations align: Universality in representation learning dynamics, 2024.
\newblock \url{https://arxiv.org/abs/2402.09142}.

\bibitem[Virtanen et~al.(2020)Virtanen, Gommers, Oliphant, Haberland, Reddy, Cournapeau, Burovski, Peterson, Weckesser, Bright, et~al.]{virtanen2020scipy}
Pauli Virtanen, Ralf Gommers, Travis~E Oliphant, Matt Haberland, Tyler Reddy, David Cournapeau, Evgeni Burovski, Pearu Peterson, Warren Weckesser, Jonathan Bright, et~al.
\newblock Scipy 1.0: fundamental algorithms for scientific computing in python.
\newblock \emph{Nature methods}, 17\penalty0 (3):\penalty0 261--272, 2020.

\bibitem[Vogelsang et~al.(2024)Vogelsang, Vogelsang, Pipa, Diamond, and Sinha]{Vogelsang2024}
Lukas Vogelsang, Marin Vogelsang, Gordon Pipa, Sidney Diamond, and Pawan Sinha.
\newblock Butterfly effects in perceptual development: A review of the ‘adaptive initial degradation’ hypothesis.
\newblock \emph{Developmental Review}, 71:\penalty0 101117, March 2024.
\newblock ISSN 0273-2297.
\newblock \doi{10.1016/j.dr.2024.101117}.

\bibitem[Yamins and DiCarlo(2016)]{Yamins2016}
Daniel L~K Yamins and James~J DiCarlo.
\newblock Using goal-driven deep learning models to understand sensory cortex.
\newblock \emph{Nature Neuroscience}, 19\penalty0 (3):\penalty0 356–365, February 2016.
\newblock ISSN 1546-1726.
\newblock \doi{10.1038/nn.4244}.
\newblock \url{http://dx.doi.org/10.1038/nn.4244}.

\bibitem[Yamins et~al.(2014{\natexlab{a}})Yamins, Hong, Cadieu, Solomon, Seibert, and DiCarlo]{Yamins2014}
Daniel L.~K. Yamins, Ha~Hong, Charles~F. Cadieu, Ethan~A. Solomon, Darren Seibert, and James~J. DiCarlo.
\newblock Performance-optimized hierarchical models predict neural responses in higher visual cortex.
\newblock \emph{Proceedings of the National Academy of Sciences}, 111\penalty0 (23):\penalty0 8619–8624, May 2014{\natexlab{a}}.
\newblock ISSN 1091-6490.
\newblock \doi{10.1073/pnas.1403112111}.
\newblock \url{http://dx.doi.org/10.1073/pnas.1403112111}.

\bibitem[Yamins et~al.(2014{\natexlab{b}})Yamins, Hong, Cadieu, Solomon, Seibert, and DiCarlo]{yamins2014performance}
Daniel~LK Yamins, Ha~Hong, Charles~F Cadieu, Ethan~A Solomon, Darren Seibert, and James~J DiCarlo.
\newblock Performance-optimized hierarchical models predict neural responses in higher visual cortex.
\newblock \emph{Proceedings of the national academy of sciences}, 111\penalty0 (23):\penalty0 8619--8624, 2014{\natexlab{b}}.

\bibitem[Zhuang et~al.(2021)Zhuang, Yan, Nayebi, Schrimpf, Frank, DiCarlo, and Yamins]{Zhuang2021}
Chengxu Zhuang, Siming Yan, Aran Nayebi, Martin Schrimpf, Michael~C. Frank, James~J. DiCarlo, and Daniel L.~K. Yamins.
\newblock Unsupervised neural network models of the ventral visual stream.
\newblock \emph{Proceedings of the National Academy of Sciences}, 118\penalty0 (3), January 2021.
\newblock ISSN 1091-6490.
\newblock \doi{10.1073/pnas.2014196118}.
\newblock \url{http://dx.doi.org/10.1073/pnas.2014196118}.

\end{thebibliography}


\newcommand{\beginsupplement}{
    \setcounter{table}{0}
    \renewcommand{\thetable}{S\arabic{table}}%
    \setcounter{figure}{0}
    \renewcommand{\thefigure}{S\arabic{figure}}%
    \setcounter{equation}{0}
    \renewcommand{\theequation}{S\arabic{equation}}%
}
\beginsupplement

\end{document}